%% file: main.tex
\pdfoutput=1
\documentclass{article}
\usepackage{iclr2027_conference,times}

\input{math_commands.tex}

\usepackage{hyperref}
\usepackage{url}
\usepackage{booktabs} 
\usepackage{array}
\usepackage{amsfonts}
\usepackage{nicefrac}
\usepackage{microtype}
\usepackage{xcolor}
\usepackage{amsmath}
\usepackage{amssymb}
\usepackage[capitalize,noabbrev]{cleveref}
\usepackage{dsfont}
\usepackage{inconsolata}
\usepackage{subcaption}
\usepackage{wrapfig}
\usepackage{multirow}
\usepackage{enumitem}
\usepackage{tikz}
\usetikzlibrary{arrows.meta, calc, shadows.blur}

\newlength{\itemizelength} 
\input{macros}

\title{Dataset inference for data provenance and \\ privacy auditing in tabular foundation models}

\author{
Dariush Wahdany\thanks{Equal Contribution. Correspondence to \texttt{adam.dziedzic@cispa.de} and \texttt{jesse@layer6.ai}}\, ${}^1$ \quad 
Jesse C. Cresswell${}^*$${}^2$ \quad
Naiqing Guan${}^2$ \quad
Atiyeh Ashari Ghomi${}^2$ \\
\ \textbf{Franziska Boenisch}${}^1$ \quad
\textbf{Adam Dziedzic}${}^1$
\\
\\
${}^1$CISPA Helmholtz Center for Information Security \quad
${}^2$Layer 6 AI
}

\iclrfinalcopy
\begin{document}

\maketitle

\begin{abstract}
\input{content/0-abstract}
\end{abstract}

\section{Introduction}
\label{sec:intro}
\vspace{-4pt}
\input{content/1-intro}

\vspace{-4pt}
\section{Background and Related Work}
\label{sec:background}
\input{content/2-background}

\input{content/3-method}

\vspace{-4pt}
\section{Experimental Evaluation}
\label{sec:results}
\vspace{-4pt}
\input{content/5-results}

\vspace{-4pt}
\section{Conclusion}
\label{sec:conclusion}
\vspace{-4pt}
\input{content/6-conclusion}

\bibliography{bib}
\bibliographystyle{iclr2027_conference}

\appendix
\newpage

\section{Limitations and Ethical Considerations}
\label{sec:app-limitations}
\input{content/D-appendix}

\section{Extended Background on Tabular Foundation Models}
\label{sec:app-a}
\input{content/A-appendix}

\newpage

\section{Dataset Inference Details}
\label{sec:app-b}
\input{content/B-appendix}

\newpage

\section{Additional Experimental Details and Results}
\label{sec:app-c}
\input{content/C-appendix}

\end{document}

%% file: math_commands.tex
\usepackage{amsmath,amsfonts,bm}

\def\eqref#1{equation~\ref{#1}}

\def\1{\bm{1}}

\DeclareMathAlphabet{\mathsfit}{\encodingdefault}{\sfdefault}{m}{sl}
\SetMathAlphabet{\mathsfit}{bold}{\encodingdefault}{\sfdefault}{bx}{n}

\DeclareMathOperator*{\argmax}{arg\,max}

%% file: macros.tex
\newcommand{\sus}{\mathcal{D}_\text{sus}}
\newcommand{\ho}{\mathcal{D}_\text{known}}

\newcommand{\pmcell}[2]{\begin{tabular}[c]{@{}c@{}}#1\\$\pm$#2\end{tabular}}

%% file: content/0-abstract.tex
\vspace{-4pt}Tabular foundation models (TFMs) are increasingly deployed through black-box APIs and trained on large collections of real-world tabular datasets. As this training paradigm becomes more widely adopted, private or proprietary datasets may be incorporated into pre-training corpora, yet there are currently no dedicated methods for determining whether a given tabular dataset was used to train a TFM. As a solution, we introduce the first dataset inference method for TFMs, aiming to infer whether a suspect dataset was part of a model's pre-training data. We systematically analyze a broad collection of candidate signals that can be observed from a black-box TFM via input manipulation and find that individual signals provide only weak evidence of training data membership. However, by combining multiple complementary signals, we can infer dataset membership for several state-of-the-art TFMs trained on real tabular data, achieving up to 0.99~ROCAUC dependent on the TFM. We then study factors that influence dataset identification, including pre-training data composition, model capacity, and the use of real vs. synthetic data. Our results show that our dataset inference method is a practical auditing tool for detecting privacy leakage and the use of proprietary datasets in TFMs.

%% file: content/1-intro.tex
Foundation models are a powerful paradigm for creating versatile models on various data modalities. As a prototype, large language models (LLMs) use transformer architectures \citep{vaswani2017attention} to process vast amounts of pre-training data, guided by scaling laws that predict performance improvements \citep{kaplan2020scaling}. In contrast to the rapid progress of this paradigm in the language domain, performance on predictive tabular data tasks has been dominated by gradient-boosted tree-based methods for most of the previous decade \citep{chen2016xgboost,ke2017lightgbm, prokhorenkova2018catboost}. Only with the rise of prior-data fitted networks \citep{muller2022transformers, hollmann2023tabpfn, Hollmann2025}, a transformer-based architecture which exhibits scaling properties similar to LLMs \citep{ma2025tabdpt}, have tabular foundation models (TFMs) begun to convincingly outperform tree-based models \citep{erickson2025tabarena}. TFMs are pre-trained on a wide variety of datasets, then for inference take a labeled dataset as context and directly predict on unlabeled examples via in-context learning \citep{brown2020language}, without requiring updates to model weights. This ease of use combined with their strong predictive power has rapidly made TFMs a go-to tool across industries where tabular data is prevalent.

One explanation for the delayed development of TFMs compared to foundation models for other modalities is the relative scarcity of large-scale and high-quality tabular datasets. Unlike text or images which are available in vast quantities on the internet, publicly available tabular datasets tend to be small. The first TFMs circumvented this issue by relying on synthetic data generated from a prior distribution over tabular datasets, which set the trend for many subsequent TFMs \citep{qu2025tabicl,qu2026tabiclv2betterfasterscalable,  grinsztajn2025tabpfn, zhang2025limix, zhang2025mitra, bouadi2025orion, hosseinzadeh2026tabdpt}. However, scaling laws for TFMs \citep{ma2025tabdpt} indicate that real-world datasets can lead to better scaling exponents, suggesting that it may be worthwhile to curate larger-scale real-world tabular datasets for pre-training. Recently, there has been increased effort from the research community in this direction \citep{vanschoren2014openml, eggert2023tablib, spinaci2025contexttab}.

Unfortunately, the drive to curate ever larger datasets for foundation model consumption has created controversy over ownership and licensing for scraped data \citep{nyt2023sue, carlini2023quantifying, freeman2024exploring, ross2025a}. In the tabular space, further pressure has been added due to the rapid commercialization of TFMs hosted as black-box APIs \citep{priorlabs_tabpfn_large_data, fundamental,kumo2025rfm, neuralk2026}. Creators of black-box TFMs have an incentive to train on as much real-world tabular data as possible, without diligently obtaining permission from the respective data owners or subjects. Yet there is currently no dedicated method for data owners to determine whether their dataset was used to train a black-box TFM without permission.

This lack of training data transparency creates an auditing problem for both model developers and external data owners.
Organizations training or fine-tuning TFMs on sensitive internal data will need to assess the risk of data leakage before releasing their models. Such auditing is especially important in domains such as finance, e-commerce, and healthcare where tabular data is prevalent and often contains highly sensitive information.  Conversely, data owners may want to verify whether their datasets were incorporated into a TFM without authorization. However, training-data identification for TFMs is challenging: the models have diverse architectures, attention mechanisms, and may be trained on datasets that are repeatedly subsampled, augmented, transformed, and redefined into different prediction tasks, in the manner of self-supervised learning \citep{chen2020simclr, balestriero2023cookbook, sui2024self, ma2025tabdpt, shaheen2026understanding}. These training procedures can substantially weaken direct evidence that a given dataset was part of pre-training.

For these reasons, we investigate the feasibility of \emph{dataset inference} (DI) for TFMs. Similar to membership inference~\citep{shokri2017membership,carlini2022membership}, which aims to determine whether an individual data point was part of a model's training set, DI asks whether an entire dataset was included in the training corpus of a model~\citep{maini2021dataset}. DI has previously been studied for self-supervised vision models~\citep{dziedzic2022dataset}, LLMs~\citep{maini2024llm,rossi2026natural,zhao2025unlocking}, and image generative models~\citep{dubinski2025cdi,kowalczuk2025privacy}. TFMs, however, differ from these settings because their predictions depend jointly on the model parameters, the provided context dataset, and the unlabeled query points. This creates both new challenges and new opportunities for DI. In this work, we systematically design TFM-specific input manipulations and output signals to create the first training dataset inference method for TFMs. 

Our main contributions are as follows:
\begin{enumerate}[itemsep=0pt, topsep=0pt]
\setlength{\leftskip}{\itemizelength}
    \item We formulate the dataset inference task for TFMs and design TFM-specific signals for identification of training datasets.
    \item We show that individual signals are weak, but that multi-signal inference can accurately identify training datasets in some state-of-the-art TFMs.
    \item We analyze factors in the models and data that drive dataset detectability,  and identify pre-training data composition as the main driver of detectability: real and real-derived synthetic datasets leave strong dataset-level traces, while model size alone has little effect and prior-synthetic data leaks only when very few datasets are used. 
    \item Our experiments highlight the potential of dataset inference as a practical auditing tool for detecting the use of private or proprietary datasets in TFMs.
\end{enumerate}
\vspace{-2pt}

%% file: content/2-background.tex
\vspace{-2pt}
\subsection{Tabular Foundation Models}
\vspace{-4pt}
\label{sec:tfms}
\begin{wrapfigure}[6]{r}{0.36\linewidth}
    \centering
    \vspace{-14pt}
    \resizebox{\linewidth}{!}{%
    \input{figures/tfm_prediction2}
    }
    \caption{\textbf{Prediction with TFMs.}}
    \label{fig:prediction}
\end{wrapfigure}
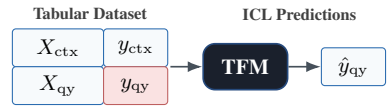
TFMs are highly diverse in their architectural details, data treatments, and training methods. The primary commonality is the in-context learning (ICL) prediction mechanism. Each TFM, denoted $f_\theta$, is designed to take in a labeled \emph{context} dataset $\mathcal{D}_\text{ctx} = \{X_\text{ctx}, y_{\text{ctx}}\}$, along with unlabeled \emph{query} data points $\mathcal{D}_\text{qy} = \{X_\text{qy}\}$ from the same marginal distribution, and then predicts the corresponding labels $y_{\text{qy}}$ in-context (without modifying the weights of $f_\theta$). This process of prediction with TFMs is depicted in \Cref{fig:prediction}. Note that TFMs are often designed to perform classification, regression, or both kinds of predictive tasks. More details on the diversity of architectures, data, and training used across TFMs are given in Appendix~\ref{sec:app-a}.

The first TFMs were pre-trained exclusively on synthetic tabular data generated from a prior over datasets \citep{hollmann2023tabpfn}. In this case, there were no concerns over the privacy or ownership of pre-training data. However, training on real data has shown better scaling exponents \citep{ma2025tabdpt}, indicating that the ceiling for performance is higher with real tabular data \citep{spinaci2025contexttab, garg2025real}, as has been the case for other modalities~\citep{kaplan2020scaling,Fan_2024_CVPR}. Additionally, organizations train TFMs on their own private datasets to tailor them to specialized domains \citep{breejen2024fine}. In these cases the concerns of private information leakage from TFMs are extremely relevant. To tackle a similar problem, \citet{carey2024dp} used LLMs to predict on tabular data and introduced a differentially private mechanism~\citep{dwork2006} to protect data in the \textit{prompt}. \citet{byun2025risk} looked at privacy risks of data \emph{generation} using foundation models, including tabular data generation. Neither work investigates pre-training dataset identification for predictive TFMs, which is the focus of our work.
\vspace{-6pt}
\subsection{Membership and Dataset Inference}
\vspace{-4pt}
Membership inference attacks (MIAs) attempt to answer the question of whether a suspect data point was used to train a given model~\citep{shokri2017membership, carlini2022membership}, and have become the de facto standard for auditing training data privacy in machine learning~\citep{steinke2023privacy, rossi2026natural, nasr2023tight, zanellabeguelin23bayesianDPEstimation}. However, it has recently been shown that for foundation models, MIAs do not significantly outperform random guessing as training datasets grow~\citep{duan2024membership, dasBlindBaselinesBeat2025, maini2024llm}. Thus, MIAs are not suitable for identifying training datasets for these models. 

\looseness=-1 As a solution, dataset inference (DI)~\citep{maini2021dataset} shifts the focus from individual data points to entire datasets. Originally proposed for resolving model ownership in supervised learning~\citep{maini2021dataset} and subsequently extended to self-supervised learning~\citep{dziedzic2022dataset}, DI has recently been applied to identify training datasets in LLMs~\citep{maini2024llm,rossi2026natural,zhao2025unlocking}. Beyond LLMs, DI has also been used for other generative models, including diffusion models~\citep{dubinski2025cdi} and image autoregressive models~\citep{kowalczuk2025privacy}. In these settings, DI typically extracts membership signals from individual examples in a suspect dataset, aggregates these signals across the dataset, and applies a statistical test to determine whether the dataset was used during training. TFMs, however, differ fundamentally from these models: they are predictive models pre-trained on collections of datasets, and their predictions are produced through an ICL mechanism. Consequently, the relevant unit of membership for TFMs is not an individual example, but an entire dataset. Because existing DI methods do not directly apply to this setting, we propose the first DI method for TFMs, designed to determine whether a given suspect dataset was used during pre-training.

%% file: figures/tfm_prediction2.tex
\begin{tikzpicture}[
        font=\footnotesize,
        >=Latex,
        title/.style={font=\scriptsize\bfseries, text=darkgray},
        cell/.style={
            draw=slate,
            line width=0.45pt,
            minimum height=0.56cm,
            inner xsep=4pt,
            rounded corners=1.5pt,
            align=center
        },
        feature/.style={cell, fill=cBG, draw=cHO},
        target/.style={cell, fill=redfill, draw=rededge},
        pred/.style={cell, fill=cBG, draw=cHO},
        model/.style={
            draw=modeledge,
            line width=0.7pt,
            rounded corners=5pt,
            fill=cInk,
            text=white,
            font=\footnotesize\bfseries,
            minimum width=1.25cm,
            minimum height=0.68cm,
            align=center
        },
        arrow/.style={->, line width=0.7pt, draw=slate}
    ]

    \definecolor{slate}{RGB}{70,80,90}
    \definecolor{greenfill}{RGB}{226,241,229}
    \definecolor{redfill}{RGB}{250,226,226}
    \definecolor{rededge}{RGB}{190,95,95}
    \definecolor{predfill}{RGB}{222,241,232}
    \definecolor{prededge}{RGB}{80,145,105}
    \definecolor{modelfill}{RGB}{72,112,168}
    \definecolor{modeledge}{RGB}{38,74,125}

    \definecolor{cHO}{HTML}{2B6CB0}
    \definecolor{cSUS}{HTML}{C53030}
    \definecolor{cAPI}{HTML}{2D3748}
    \definecolor{cBG}{HTML}{F7FAFC}
    \definecolor{cAccent}{HTML}{B7791F}
    \definecolor{cInk}{HTML}{1A202C}

    \def\xw{1.35}
    \def\yw{0.95}
    \def\h{0.56}

    \node[title] at ({(\xw+\yw)/2}, 0.50) {Tabular Dataset};
    \node[title] at ({\xw+\yw+1.95}, 0.50) {ICL Predictions};

    \node[feature, minimum width=\xw cm] (xctx)
        at ({\xw/2}, 0) {$X_{\mathrm{ctx}}$};
    \node[feature, minimum width=\yw cm] (yctx)
        at ({\xw+\yw/2}, 0) {$y_{\mathrm{ctx}}$};

    \node[feature, minimum width=\xw cm] (xqy)
        at ({\xw/2}, -\h) {$X_{\mathrm{qy}}$};
    \node[target, minimum width=\yw cm] (yqy)
        at ({\xw+\yw/2}, -\h) {$y_{\mathrm{qy}}$};

    \node[model] (tfm) at ({\xw+\yw+1.15}, {-\h/2}) {TFM};

    \node[pred, minimum width=0.95cm, minimum height=0.56cm]
        (out) at ({\xw+\yw+2.75}, {-\h/2}) {$\hat{y}_{\mathrm{qy}}$};

    \draw[arrow] ({\xw+\yw+0.04}, {-\h/2}) -- (tfm.west);
    \draw[arrow] (tfm.east) -- (out.west);

\end{tikzpicture}%

%% file: content/3-method.tex
\vspace{-8pt}
\section{Problem Setup}\label{sec:setup}
\vspace{-6pt}
Our work targets the important question: \textit{Was a given dataset used to train a TFM?} This question is highly relevant both for external and internal audits. These setups, while conceptually similar, differ in their motivation and assumptions on what knowledge the respective parties have.

\vspace{-4pt}\textbf{External Audits - Data Provenance.} External parties who own tabular datasets and suspect that these datasets were used to train a TFM without their consent would like to verify this suspicion. They only have black-box access to the TFM, which is deployed behind an API. They hold the dataset that they suspect was used to train the TFM, and some additional datasets known to be non-members, which could be unreleased private datasets or randomly generated datasets from a prior, as are commonly used to train TFMs~\citep{hollmann2023tabpfn}.

\vspace{-4pt}\textbf{Internal Audits - Privacy Leakage.} Organizations train or fine-tune TFMs on their own proprietary datasets. TFMs are most commonly used in domains where tabular data is abundant, including finance, e-commerce, and healthcare and these domains also happen to involve highly sensitive data. Therefore, it is important to internally audit the potential leakage of sensitive information from a TFM before publicly deploying it. An internal auditor needs to verify how much private information could be leaked from their hosted black-box API to an outside user in the worst case. For this setting, we assume the auditor only accesses the black-box API, but to mimic a worst case they assume knowledge of the pre-training dataset members, and also have non-member datasets which could be, for example, internal test datasets used to assess the performance of the TFM, or synthetic data.

We consider dataset inference for both external and internal audits, where the external setting is more challenging as it assumes less knowledge about the TFM's pre-training datasets.

\vspace{-8pt}
\section{Dataset Inference for TFMs}
\label{sec:method}
\vspace{-4pt}
Our DI method for TFMs aims to answer the question of whether a labeled \emph{suspect} dataset $\sus$ was used to train $f_\theta$. Therefore, following prior work on DI~\citep{maini2021dataset,dziedzic2022dataset,maini2024llm}, it utilizes the trained model $f_\theta$ and additional \emph{known} datasets $\{\ho\}$. For external audits, $\ho$ are all non-members, whereas for internal audits $\ho$ are a mix of members and non-members. Both internal and external audits use only black-box access to $f_\theta$, as through an API. This means the auditor can control the inputs to $f_\theta$, namely $ X^i_\text{qy}$ and $\mathcal{D}_\text{ctx}$, and receives the TFM's output distribution $p(y^i_\text{qy} \mid X^i_\text{qy}; \mathcal{D}_\text{ctx})$.

Using $f_\theta$, the auditor extracts a single fingerprint vector $\bar{S}(\mathcal{D}) \in \mathbb{R}^d$ from each dataset $\mathcal{D}$, applied both to the suspect dataset $\sus$ and to the known datasets $\{\ho\}$. Compared to DI in the language~\citep{maini2024llm} or vision domains~\citep{dubinski2025cdi} which calculates membership signals \textit{per sample} and then aggregates over $\sus$, in TFMs $\sus$ is treated as \textit{a single unit}. This is because TFM pre-training performs batching at the dataset level. Given $\bar{S}(\sus)$ and $\{\bar{S}(\ho)\}$, the auditor leverages these fingerprints to determine whether $\sus$ was used for training. The concrete procedure depends on the auditing setup, i.e., internal vs.\ external audit, and we instantiate both in the following sections. 

Constructing $\bar{S}(\mathcal{D})$ is the central challenge. TFMs have diverse architectures, attention mechanisms, and augment their pre-training datasets in various ways. These factors disrupt direct membership signals that could be gathered from $\sus$, and mean that different signals can be informative or not based on the TFM. Below, we show how to carefully craft a wide variety of membership signals for TFMs that reflect the diversity in their pre-training, then turn to the two auditing setups.

\vspace{-6pt}
\subsection{Crafting TFM-Specific Dataset Membership Feature Vectors}
\label{subsec:signals}
\vspace{-4pt}
\begin{wrapfigure}[16]{r}{0.4\linewidth}
  \centering
  \includegraphics[width=0.8\linewidth]{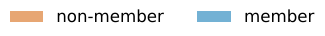}\\[-0.2em]
  \includegraphics[height=1.4in]{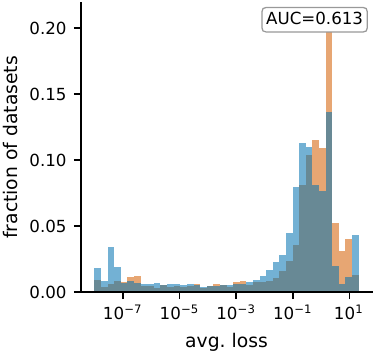}
  \caption{\textbf{Classical inference signals are weak predictors.} \textit{Per-dataset} avg. loss does not clearly separate members and non-members of SAP-RPT-1-OSS.
  }
  \label{fig:signal-separation-single}
  \vspace{-15pt}
\end{wrapfigure}
A natural starting point is to port standard membership signals, like \textit{loss} or \textit{predictive confidence}~\citep{shokri2017membership, carlini2022membership}, that have shown success in other domains to TFM DI. As \Cref{fig:signal-separation-single} shows for the loss, these signals on their own do not separate member from non-member datasets in TFMs. Since TFMs may see millions of datasets during pre-training, sometimes under many subsamplings, shuffles, and task redefinitions, any single readout captures only a faint trace of membership.

Instead, we take full advantage of the powerful lever an auditor has, namely complete control over the inputs $ X^i_\text{qy}$ and $\mathcal{D}_\text{ctx}$. These inputs can be created from the dataset $\mathcal{D}$ (either $\sus$ or one of the $\ho$) in a wide variety of ways, complementing the great diversity in TFM design. We therefore probe models using inputs crafted with specialized TFM augmentations, and assemble the responses into the fingerprint vector $\bar{S}(\mathcal{D}) \in \mathbb{R}^d$.

Each entry of $\bar{S}(\mathcal{D})$ is generated from a triple (manipulation, signal, reduction): a \emph{manipulation} perturbs the TFM's inputs; a \emph{signal} is a scalar read from the TFM's output; a \emph{reduction} first varies the manipulation's parameters, then collapses the resulting sequence of signals into a single number. Stacking the reduction values across all valid triples yields the fingerprint $\bar{S}(\mathcal{D})$. We describe each of these axes now, starting from the outputs.

\textbf{Signals} are scalar functions of the TFM's output distribution $p(y^i_\text{qy} \mid X^i_\text{qy}; \mathcal{D}_\text{ctx})$, and are aggregated over the query points $X^i_\text{qy}$. Many classical MIA signals (correct-class confidence, max confidence, margin, loss) are useful in this context.  
However, since the auditor does not know details of the TFM architecture or training methodology, they cannot assume particular details such as the exact loss that $f_\theta$ was trained with. Hence, our signals cover a variety of quantities, including common losses. The full list of 14 signals we use is given in \Cref{tab:signals} in Appendix~\ref{sec:app-b}.

\textbf{Manipulations} perturb the TFM's inputs $X^i_\text{qy}$ and $\mathcal{D}_\text{ctx}$ in structured ways that take advantage of the common ICL prediction mechanism used by TFMs. Manipulations are designed so that they will generate different patterns of signals when applied to member or non-member datasets. Furthermore, each manipulation comes with parameters that can be adjusted, such as the fraction of $\mathcal{D}$ to split into query vs. context. The general idea is that when $\mathcal{D}$ is a member, $f_\theta$ may have internalized some parametric knowledge about the relationship between $X$ and $y$ that allows it to predict more correctly or confidently on queries. While all 20 manipulations we use are detailed in \Cref{tab:manipulations} in Appendix~\ref{sec:app-b}, here we give select examples to build intuition:
\begin{itemize}[itemsep=1pt, topsep=1pt]\setlength{\leftskip}{\itemizelength}
    \item \textbf{M3 Query only}: This manipulation splits $\mathcal{D}$, but then discards $\mathcal{D}_\text{ctx}$ and inputs only $X_\text{qy}$ to $f_\theta$. If $\mathcal{D}$ is a member, the TFM will have encountered inputs like $X_\text{qy}$ before and may predict some labels correctly, whereas for non-members predictions are more likely to be random.
    \item \textbf{M17 Brainwash}: This manipulation selects rows of $\mathcal{D}_\text{qy}$ and copies them into $\mathcal{D}_\text{ctx}$ with corrupted labels $y_\text{ctx}$. If $\mathcal{D}$ is a non-member $f_\theta$ can only rely on the context and will likely predict $y_\text{qy}$ incorrectly, but for a member dataset $f_\theta$ retains parametric knowledge that can overcome the corrupted context to make correct predictions. This is inspired by Brainwash~\citep{wen2024brainwash}.
\end{itemize}

\textbf{Reductions} harness the various parameters of manipulations to create a sequence of signals which are then converted into a single scalar feature for $\bar{S}(\mathcal{D})$. We use a library of 11 reductions organised by the structure of the manipulation's parameters, given in \Cref{tab:reductions} of Appendix~\ref{sec:app-b}. When parameters correspond to a numeric intensity axis (e.g.\ noise scale, context size), the sequence forms a curve and we reduce based on its shape, for example taking its \textit{slope} or the \textit{area under the curve}. When parameters lack a directionality (e.g.\ permutations), we can still reduce with statistical metrics such as \textit{mean}, \textit{standard deviation}, and \textit{range}. Finally, for manipulations that only affect the context, predictions on the same query row are directly comparable across parameter settings, and we add cross-parameter agreement scores like \textit{consistency} and \textit{rank correlation} to capture prediction stability.

In summary, we form a large number of (manipulation, signal, reduction) triples by taking many combinations from the three sets of options ($678$ in total). The resulting reduction values are the features in the fingerprint $\bar{S}(\mathcal{D})$ used for DI. As an example, the feature \texttt{brainwash\_loss\_slope} corresponds to the Brainwash manipulation, measuring the loss signal as the Brainwash parameter (fraction of duplicated rows) is changed, and then computes the slope of that signal curve in parameter space to give a value which can be added to $\bar{S}(\mathcal{D})$.

\begin{figure}[t]
    \centering
    \resizebox{0.88\linewidth}{!}
    {\input{figures/fig_overview_2.tex}}
    \vspace{-4pt}
    \caption{\textbf{Overview of our DI procedure}. The auditor queries the black-box TFM API with both known datasets $\{\ho\}$ and a suspect dataset $\sus$ under structured input manipulations, yielding fingerprint vectors $\{\bar{S}(\ho)\}$ and $\bar{S}(\sus)$. \textit{(a) External audit:} All $\ho$ are non-members. We fit an Isolation Forest on $\{\bar{S}(\ho)\}$ and score $\bar{S}(\sus)$ as a candidate anomaly indicating membership. \textit{(b) Internal audit:} $\{\ho\}$ has a mix of member and non-members. We train a binary membership classifier $g_\varphi$ on $\{\bar{S}(\ho)\}$, and predict the membership label of $\bar{S}(\sus)$.
    }
    \label{fig:overview}
    \vspace{-14pt}
\end{figure}

\vspace{-6pt}
\subsection{External Auditing}
\label{subsec:unsupervised}
\vspace{-4pt}
For external auditing, the auditor, i.e., the data owner, has black-box TFM access and known non-member datasets $\{\ho\}$, along with their suspect dataset $\sus$. Since the auditor only knows \textit{non-member} datasets and does not already know any \textit{members}, we frame DI as anomaly detection calibrated only on non-members as depicted in \Cref{fig:overview}. We fit an Isolation Forest~\citep{liu2008isolation,liu2012isolation} on the known datasets' fingerprints $\{\bar{S}(\ho)\}$ to characterize the typical non-member behaviour, and score the suspect dataset $\bar{S}(\sus)$ against this baseline. If the suspect falls outside the normality region, it is flagged as likely to be a pre-training member. This formulation requires neither examples of member datasets, nor training expensive shadow models \citep{shokri2017membership}.

Retaining all available features in $\bar S$ would create unnecessary noise due to features that measure distribution shifts within non-members, rather than membership information. To ensure each anomaly detection feature is meaningful we randomly split $\{\bar{S}(\ho)\}$ in halves, and as a baseline measure the separation a feature produces. Then, that feature is retained only if its suspect-versus-known separation exceeds the 95th percentile of the baseline.

To evaluate the method, we score a set of $\{\sus\}$ which contains the member datasets and half of the non-member datasets, none of which are part of $\{\ho\}$. Given the Isolation Forest scores $S_{\text{known}} = \{s_i\}_{i=1}^{n}$ for $\{\ho\}$ we turn an individual score $s_\text{sus}$ for a dataset $\sus$ into a conformal $p$-value~\citep{vovk2005algorithmic, bates2023outliers}, i.e., 
$p(s_\text{sus}) = \big(1 + \textstyle\sum_{i=1}^{n} \mathds{1}[s_i \geq s_\text{sus}]\big) / (n+1)$ which can be thresholded.
\vspace{-4pt}
\subsection{Internal Auditing}
\label{subsec:supervised}
\vspace{-4pt}
In addition to black-box access an internal auditor also has ground truth knowledge of which datasets were used for training the TFM. This follows the setup from prior DI works~\citep{maini2021dataset, maini2024llm,kowalczuk2025privacy,dubinski2025cdi}. Therefore, unlike for the external auditor, $\{\ho\}$ contains non-member and member datasets. For the sake of testing, the auditor removes one dataset from $\{\ho\}$ to play the role of $\sus$, ignores its membership label, and tries to determine the label from other available information. We evaluate the DI method holistically by applying leave-one-out cross-validation (LOOCV), in which we iterate the choice of $\sus$ over the whole set $\{\ho\}$. In each iteration, given $\{\ho\}$ with their membership labels $y_i \in \{0, 1\}$ (non-member vs.\ member) and fingerprints $\{\bar{S}(\ho)\}$, we train a binary classifier $g_\varphi$ to predict $y_i$, and apply it to $\bar{S}(\sus)$, as depicted in \Cref{fig:overview}. We then report the ROCAUC of the left-out predictions over the LOOCV setup. 

Again, for internal auditing we filter out features from $\bar S$ to reduce noise. Within each LOOCV fold we use a validation split to rank features by mean absolute SHAP value \citep{lundberg2017shap, lundberg2019consistentindividualizedfeatureattribution} based on a fit to $\bar S$ and the membership labels of that fold's training datasets, using a gradient-boosted tree for $g_\varphi$ \citep{chen2016xgboost}. Then we retain only the top $K$ features, with $K$ maximizing the validation ROCAUC divided by $\log_2(K+1)$.

\vspace{-4pt}
\subsection{Blind Baselines}
\vspace{-4pt}
As we cannot assume that the member and non-member dataset follow the same distribution we employ \textit{blind baselines}~\citep{dasBlindBaselinesBeat2025} when evaluating both external and internal auditing power. If member and non-member datasets come from different distributions, it may be possible for an auditing procedure to distinguish them purely from distributional differences, not from membership information revealed by a TFM. This would constitute a false positive for identifying information leakage from a TFM. Hence, the blind baseline's purpose is to differentiate signals due to distribution shifts from signals induced by training on the data.

For this purpose we replace the target TFM with a ``blind'' TFM that has entirely different pre-training members, but we use the same datasets $\{\ho\}$ for calibration, and $\{\sus\}$ for testing without changing their membership labels. Since none of $\{\sus\}$ are true members of the blind TFM, any observed differences between predictions on $\{\sus\}$ and $\{\ho\}$ must be attributed to the data distribution itself, not membership leakage from a TFM. The excess accuracy on the target TFM over the blind baseline indicates a true DI signal.

%% file: figures/fig_overview_2.tex
\definecolor{cHO}{HTML}{2B6CB0}
\definecolor{cSUS}{HTML}{C53030}
\definecolor{cAPI}{HTML}{2D3748}
\definecolor{cBG}{HTML}{F7FAFC}
\definecolor{cAccent}{HTML}{B7791F}
\definecolor{cInk}{HTML}{1A202C}

\begin{tikzpicture}[
    font=\sffamily\small,
    >={Stealth[length=4.5pt,width=4pt]},
    line cap=round,
    line join=round,
]

\def\sxA{0}
\def\sxB{3.6}
\def\sxC{7.0}
\def\sxD{11.0}
\def\sxE{14.7}

\tikzset{
    panel/.style={rounded corners=3pt, draw=cInk!15, line width=0.4pt, fill=cBG, inner sep=8pt},
    arr/.style={-{Stealth[length=4.5pt,width=4pt]}, line width=0.7pt, draw=cInk!55},
    arrHO/.style={-{Stealth[length=4.5pt,width=4pt]}, line width=0.8pt, draw=cHO!75},
    arrSUS/.style={-{Stealth[length=4.5pt,width=4pt]}, line width=0.8pt, draw=cSUS!75},
}

\begin{scope}[shift={(\sxA, 0.75)}]
    \draw[draw=cHO!45, fill=cHO!4, rounded corners=1.5pt, line width=0.4pt]
        (-1.05, -0.18) rectangle (0.65, 0.50);
    \draw[draw=cHO!60, fill=cHO!7, rounded corners=1.5pt, line width=0.45pt]
        (-0.95, -0.26) rectangle (0.75, 0.42);
    \draw[draw=cHO, fill=cHO!10, rounded corners=1.5pt, line width=0.5pt]
        (-0.85, -0.34) rectangle (0.85, 0.34);
    \foreach \y in {-0.17, 0, 0.17} \draw[cHO!30, line width=0.25pt] (-0.85,\y) -- (0.85,\y);
    \draw[cHO!30, line width=0.25pt] (0.45,-0.34) -- (0.45,0.34);
    \node[font=\bfseries, text=cHO!70!black] at (0, 0.72) {$\{\ho\}$};
    \node[font=\scriptsize\itshape, text=cHO!60!black] at (0,-0.58) {membership labels known};
\end{scope}

\begin{scope}[shift={(\sxA, -1.10)}]
    \draw[draw=cSUS, fill=cSUS!10, rounded corners=1.5pt, line width=0.5pt]
        (-0.85,-0.34) rectangle (0.85,0.34);
    \foreach \y in {-0.17, 0, 0.17} \draw[cSUS!30, line width=0.25pt] (-0.85,\y) -- (0.85,\y);
    \draw[cSUS!30, line width=0.25pt] (0.45,-0.34) -- (0.45,0.34);
    \node[font=\bfseries, text=cSUS!70!black] at (0, 0.62) {$\sus$};
    \node[font=\scriptsize\itshape, text=cSUS!60!black] at (0,-0.58) {membership label unknown};
\end{scope}

\node[draw=cAPI, line width=0.7pt, rounded corners=4pt,
      fill=cAPI, minimum width=2.2cm, minimum height=2.6cm,
      drop shadow={shadow xshift=0.6pt, shadow yshift=-0.6pt, opacity=0.18}]
      (api) at (\sxB, -0.25) {};
\node[font=\bfseries, text=white] at (api.center) {TFM};
\node[font=\tiny, text=white!80] at ($(api.center)+(0,-0.36)$) {(black-box API)};
\fill[cHO!85] ($(api.west)+(0,0.5)$) circle (0.07);
\fill[cSUS!85] ($(api.west)+(0,-0.5)$) circle (0.07);
\fill[white!70] ($(api.east)+(0,0.5)$) circle (0.07);
\fill[white!70] ($(api.east)+(0,-0.5)$) circle (0.07);

\draw[arrHO] (\sxA+0.85, 0.75) -- ($(api.west)+(0,0.5)$);
\draw[arrSUS] (\sxA+0.85, -1.10) -- ($(api.west)+(0,-0.5)$);

\newcommand{\sigmatrix}[3]{%
    \begin{scope}[shift={(#2,#3)}]
        \foreach \r/\rowdata in {0/{30,55,15,80,40,65,10,45},
                                  1/{60,25,75,35,50,80,20,55},
                                  2/{50,40,65,25,15,55,75,35}} {
            \foreach \c [count=\ci from 0] in \rowdata {
                \fill[#1!\c!white] ({\ci*0.17-0.68}, {\r*0.17}) rectangle ({\ci*0.17+0.15-0.68}, {\r*0.17+0.15});
            }
        }
        \draw[#1!55, rounded corners=1pt, line width=0.4pt] (-0.705,-0.025) rectangle (0.705,0.655);
    \end{scope}
}

\begin{scope}[shift={(\sxC, 0.435)}]
    \draw[draw=cHO!45, fill=cHO!4, rounded corners=1pt, line width=0.4pt]
        (-0.905, 0.175) rectangle (0.505, 0.855);
    \draw[draw=cHO!60, fill=cHO!7, rounded corners=1pt, line width=0.45pt]
        (-0.805, 0.075) rectangle (0.605, 0.755);
\end{scope}
\sigmatrix{cHO}{\sxC}{0.435}
\node[font=\bfseries\footnotesize, text=cHO!70!black] at (\sxC, 1.45) {$\{\bar{S}(\ho)\}$};

\sigmatrix{cSUS}{\sxC}{-1.415}
\node[font=\bfseries\footnotesize, text=cSUS!70!black] at (\sxC, -0.40) {$\bar{S}(\sus)$};

\draw[arrHO] ($(api.east)+(0,0.5)$) -- (\sxC-0.7, 0.75);
\draw[arrSUS] ($(api.east)+(0,-0.5)$) -- (\sxC-0.7, -1.10);

\begin{scope}[shift={(\sxD, -1.05)}]
    \fill[cBG, rounded corners=3pt] (-1.5,-0.65) rectangle (1.5,0.65);
    \draw[cInk!12, rounded corners=3pt, line width=0.4pt] (-1.5,-0.65) rectangle (1.5,0.65);

    \draw[cInk!50, dashed, line width=0.6pt] (-1.30, 0.55) -- (1.30, -0.55);
    \node[
    font=\bfseries\small,
    text=cInk!75,
    fill=white,
    inner sep=1pt
] at (0.48,-0.4) {$g_\varphi$};

    \foreach \x/\y in {
        0.30/0.45, 0.55/0.20, 0.50/0.50, 0.75/0.05, 0.85/0.40,
        0.95/0.50, 0.20/0.50, 1.05/0.25, 0.65/0.40, 1.15/0.45,
        0.40/0.30, 0.78/0.25,
        -0.65/-0.25, -0.85/0.00, -0.50/-0.40, -1.05/-0.15, -0.35/-0.50,
        -0.75/0.15, -1.15/0.10, -0.25/-0.35, -0.55/-0.05, -0.95/-0.40,
        -0.45/-0.20
    } {
        \fill[cHO] (\x,\y) circle (0.04);
    }

    \foreach \x/\y in {-0.20/-0.15} {
        \draw[white, line width=3.6pt, line cap=round]
            (\x-0.09,\y-0.09) -- (\x+0.09,\y+0.09)
            (\x-0.09,\y+0.09) -- (\x+0.09,\y-0.09);
        \draw[cSUS, line width=1.6pt, line cap=round]
            (\x-0.09,\y-0.09) -- (\x+0.09,\y+0.09)
            (\x-0.09,\y+0.09) -- (\x+0.09,\y-0.09);
    }

\end{scope}
\begin{scope}[shift={(\sxD, 0.95)}]
    \fill[cBG, rounded corners=3pt] (-1.5,-0.85) rectangle (1.5,0.85);
    \draw[cInk!12, rounded corners=3pt, line width=0.4pt] (-1.5,-0.85) rectangle (1.5,0.85);

    \foreach \r/\op in {0.78/6, 0.62/14, 0.44/24, 0.26/36} {
        \fill[cHO!\op] (0,0) circle (\r);
    }
    \draw[cHO!55, dashed, line width=0.35pt] (0,0) circle (0.62);

    \foreach \x/\y in {0.08/0.03, -0.15/0.13, 0.02/-0.08, 0.18/0.02, -0.22/-0.03,
                       -0.04/0.13, 0.15/-0.15, -0.16/-0.11, 0.23/0.10, -0.10/-0.18,
                       0.04/0.18, 0.26/-0.04, -0.24/0.06, 0.10/0.21, -0.08/0.03} {
        \fill[cHO] (\x,\y) circle (0.04);
    }

    \foreach \x/\y in {0.78/0.50} {
        \draw[white, line width=3.6pt, line cap=round]
            (\x-0.10,\y-0.10) -- (\x+0.10,\y+0.10)
            (\x-0.10,\y+0.10) -- (\x+0.10,\y-0.10);
        \draw[cSUS, line width=1.7pt, line cap=round]
            (\x-0.10,\y-0.10) -- (\x+0.10,\y+0.10)
            (\x-0.10,\y+0.10) -- (\x+0.10,\y-0.10);
    }
\end{scope}

\node[font=\bfseries\footnotesize, text=cInk!85, anchor=west] at (\sxD-1.50, 1.95)
    {4a.\ External: Isolation Forest};
\node[font=\bfseries\footnotesize, text=cInk!85, anchor=west] at (\sxD-1.50, -0.2)
    {4b.\ Internal: LOO classifier};

\draw[arrHO]
    (\sxC+0.7, 0.75) .. controls (\sxC+1.7, 0.75) and (\sxD-1.95, 0.95) .. (\sxD-1.50, 0.95);
\draw[arrSUS]
    (\sxC+0.7, -1.10) .. controls (\sxC+1.6, -1.10) and (\sxD-1.95, 0.55) .. (\sxD-1.50, 0.85);
\draw[arrHO]
    (\sxC+0.7, 0.75) .. controls (\sxC+1.6, 0.75) and (\sxD-1.95, -0.85) .. (\sxD-1.50, -1.05);
\draw[arrSUS]
    (\sxC+0.7, -1.10) .. controls (\sxC+1.7, -1.10) and (\sxD-1.95, -1.20) .. (\sxD-1.50, -1.20);

\foreach \x/\n/\t in {\sxA/1/{Inputs},
                      \sxB/2/{TFM querying},
                      \sxC/3/{Signal extraction}} {
    \node[font=\bfseries\footnotesize, text=cInk!85] at (\x, 1.85) {\n.\ \t};
}

\end{tikzpicture}

%% file: content/5-results.tex
\subsection{Experimental Setup}
\vspace{-4pt}
We present DI experiments on four TFMs: SAP-RPT-1-OSS \citep{spinaci2025contexttab}, TabDPT \citep{ma2025tabdpt}, nanoTabPFN \citep{pfefferle2025nanotabpfn}, and Real-TabPFN-2.5 \citep{grinsztajn2025tabpfn}. For SAP-RPT-1-OSS, TabDPT, and Real-TabPFN-2.5, we use the pre-trained model weights released by their authors. For nanoTabPFN we train the model using default settings. These four TFMs were each trained on a fixed corpus of datasets which define the members.\footnote{Note that the four TFMs have \emph{different} fixed corpora. Hence experimental results are not directly comparable.} Any dataset not used for pre-training a given TFM is a valid non-member, which can include held-out test datasets, and synthetic datasets generated from a prior. Details on the member and non-member datasets actually used for experiments with each model are fully described in Appendix~\ref{sec:tfms-datasets}. Notably, SAP-RPT-1-OSS and TabDPT were trained entirely on real tabular datasets, whereas nanoTabPFN was trained on a fixed set of synthetic datasets. Real-TabPFN-2.5 was pre-trained on synthetic data drawn fresh from a prior every batch and then fine-tuned on a small set of real tables. We use the real tables as members.

As a blind baseline~\citep{dasBlindBaselinesBeat2025}, we employ TabPFN-2.5 \citep{grinsztajn2025tabpfn}, which was not trained with a fixed corpus, but on datasets drawn from a prior for each batch. Hence, TabPFN-2.5 does not share any member datasets in common with the above four TFMs.
For internal auditing, we experiment with three tabular algorithms for the membership classifier $g_\varphi$: logistic regression, XGBoost~\citep{chen2016xgboost}, and TabPFN-2.5.\footnote{We use a TFM for $g_\varphi$ because TFMs are the strongest available classifiers. We do not wish to confuse the TFM $g_\varphi$ with the target TFM $f_\theta$ that is being audited or the blind baseline; they play independent roles. When $g_\varphi$ is a TFM, it is not actually trained; instead it uses $\{\bar S(\ho)\}$ as context and $\bar S(\sus)$ as query.} 

\begin{figure}[t]
    \centering
    \begin{minipage}{0.555\linewidth}
        \centering
        \includegraphics[width=0.9\linewidth]{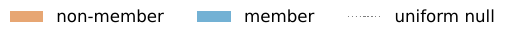}\\[-0.2em]
        \begin{subfigure}[b]{0.49\linewidth}
            \centering
            \includegraphics[height=1.2in]{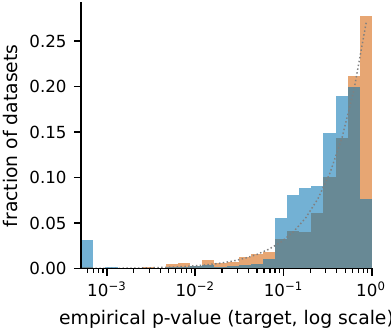}
            \caption{Target TFM}
            \label{subfig:pvalue-target}
        \end{subfigure}\hfill
        \begin{subfigure}[b]{0.49\linewidth}
            \centering
            \includegraphics[height=1.2in]{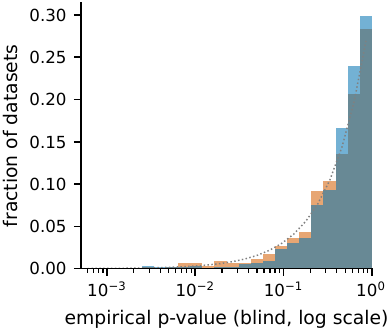}
            \caption{Blind TFM}
            \label{subfig:pvalue-blind}
        \end{subfigure}
        \caption{\textbf{External audit DI.} Empirical $p$-values from the Isolation Forest fit on $\{\bar{S}(\ho)\}$, and evaluated on various $\sus$. (\subref{subfig:pvalue-target}) shows the target TFM (SAP-RPT-1-OSS) and (\subref{subfig:pvalue-blind}) shows the blind baseline. On the target, member suspect datasets shift to lower $p$-values, with some pinned at the smallest attainable empirical $p$-value.}
        \label{fig:pvalue-sap}
    \end{minipage}\hfill
    \begin{minipage}{0.405\linewidth}
        \centering
        \setlength{\abovecaptionskip}{0pt}
        \setlength{\belowcaptionskip}{7pt}
        \captionof{table}{\textbf{External-audit DI AUC.} For each target TFM, we show the ROCAUC of membership detection by the Isolation Forest fit on $\{\bar{S}(\ho)\}$. The \emph{Blind} column reports the same pipeline run against TabPFN-2.5 and excess of target over blind quantifies the actual DI signal.}
        \label{tab:external-auc}
        \vspace{4pt}
        {\footnotesize
        \setlength{\tabcolsep}{3pt}
        \begin{tabular}{lccc}
    \toprule
        Target TFM & Target & Blind & Excess \\
        \midrule
        SAP-RPT-1-OSS & \pmcell{0.652}{0.041} & \pmcell{0.442}{0.013} & \pmcell{0.210}{0.045} \\
        TabDPT & \pmcell{0.583}{0.060} & \pmcell{0.472}{0.054} & \pmcell{0.110}{0.043} \\
        nanoTabPFN & \pmcell{0.502}{0.011} & \pmcell{0.502}{0.003} & \pmcell{$-$0.000}{0.010} \\
        Real-TabPFN-2.5 & \pmcell{0.566}{0.036} & \pmcell{0.581}{0.047} & \pmcell{$-$0.015}{0.020} \\
        \bottomrule
    \end{tabular}}
    \end{minipage}
    \vspace{-10pt}
\end{figure}

Our implementation of the methods is available at \href{https://github.com/layer6ai-labs/tfm-di}{\texttt{github.com/layer6ai-labs/tfm-di}}.

\subsection{Dataset Inference Performance}\label{sec:di-performance}
\textbf{External Auditing.} \Cref{fig:pvalue-sap} shows empirical $p$-values produced by our anomaly detection method (\Cref{subsec:unsupervised}) on fingerprints from SAP-RPT-1-OSS. On the target TFM, the distribution of anomaly scores differs significantly based on whether a given $\sus$ is a true member or not. Many member dataset fingerprints have anomaly scores that sit in the low-$p$ tail and can be confidently inferred as members, while the non-member distribution closely follows the uniform reference (dotted). On the blind TFM, both member and non-member $p$-values pile near $1$, exactly as expected when the TFM has seen neither $\sus$ nor $\ho$ during pre-training. 

\Cref{tab:external-auc} reports the corresponding ROCAUC values across all target TFMs alongside their blind baselines. On SAP-RPT-1-OSS, there is a clear signal that the DI method captures membership information from the target TFM in excess of what a blind baseline achieves. However, this excess is much weaker for TabDPT, showing that it leaks less membership information. NanoTabPFN and Real-TabPFN-2.5 do not leak any membership information under this protocol. There are of course two possible explanations: the auditing procedure does not have enough discriminatory power to extract the membership information; or, these TFMs do not retain any significant membership information in the first place. We investigate the differences between TFMs more in \Cref{sec:success_factors}.

\begin{figure}[t]
    \centering
    \begin{minipage}[t]{0.45\linewidth}
        \centering
        \includegraphics[width=0.6\linewidth]{figures/fig_signal_separation_loss_legend_2col.pdf}
        \includegraphics[height=1.2in]{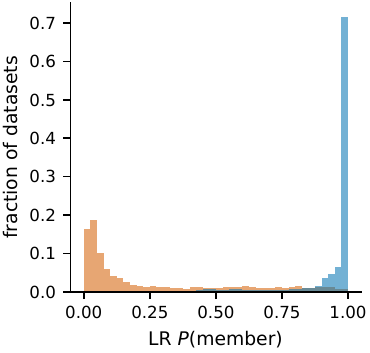}
        \captionof{figure}{\textbf{Multiple signals together succeed.} Our method (\Cref{subsec:supervised}) with multiple TFM-specific signals leads to clear separation, where single signals did not (cf. \Cref{fig:signal-separation-single}).}
        \label{fig:signal_separation}
    \end{minipage}\hfill
    \begin{minipage}[t]{0.52\linewidth}
        \centering
        \includegraphics[width=0.72\linewidth]{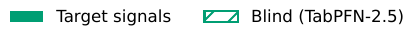}
        \includegraphics[height=1.15in]{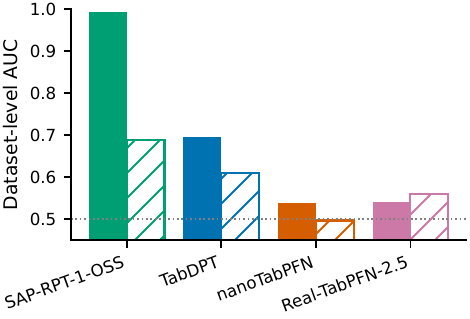}
        \vspace{-4pt}
        \captionof{figure}{\textbf{Internal-audit DI ROCAUC.} DI success varies strongly between target TFMs, from no power (Real-TabPFN-2.5) to an almost perfect ROCAUC of $0.99$ (SAP-RPT-1-OSS). Excess over the blind baseline indicates membership leakage.}
        \label{fig:attack_dia_auc}
    \end{minipage}
        \vspace{-8pt}
\end{figure}

\textbf{Internal Auditing.} \Cref{fig:signal_separation} demonstrates the positive effect of using many TFM-specialized membership features in the fingerprint $\bar S(\mathcal{D})$, as compared to \Cref{fig:signal-separation-single} which used single classic membership signals. In this demonstration we show $20{,}100$ SAP-RPT-1-OSS member and non-member datasets scored by the logistic regression classifier $g_\varphi$, as described in \Cref{subsec:supervised}. Using TFM-specialized features greatly increases the predictive power of the audit. \Cref{fig:signal-separation-tabdpt} in Appendix~\ref{sec:app-signal-separation-tabdpt} shows the same comparison for TabDPT.

\Cref{fig:attack_dia_auc} reports the internal audit ROCAUC across target TFMs using the TabPFN-2.5 classifier $g_\varphi$ compared to the blind TFM. \Cref{tab:internal-auc} shows the same information across all classifiers. We find that the classifiers $g_\varphi$ acting on fingerprints from blind baselines can achieve non-trivial performance, which highlights the necessity of accounting for distributional differences between $\sus$ and $\{ \ho\}$. Hence, we focus on the excess as true membership leakage. Success varies substantially: nanoTabPFN and Real-TabPFN-2.5 sit only just above random (AUC $\sim0.48-0.59$) for both the target TFM and the blind baseline, suggesting little to no leakage. In contrast, SAP-RPT-1-OSS reaches significantly higher discrimination than the blind baseline (e.g. AUC $0.99$ vs. $0.69$), as does TabDPT (e.g. AUC $0.70$ vs. $0.61$) indicating genuine membership leakage.

These results demonstrate for the first time that (\emph{i}) some TFMs internalize membership information about which datasets they are pre-trained on; and (\emph{ii}) our dataset inference methods can extract that information in the external and internal audit settings.
\begin{table}[t]
    \centering
    \small
    \setlength{\tabcolsep}{6pt}
    \setlength{\extrarowheight}{-1pt}
    \setlength{\aboverulesep}{0.5ex}
    \setlength{\belowrulesep}{0.5ex}
    \setlength{\cmidrulesep}{0.3ex}
    \caption{\textbf{Internal-audit DI AUC by classifier.} For each target TFM and classifier, we report the test ROCAUC of dataset-membership prediction; the \emph{Blind TFM} column reports the same pipeline run against fingerprints from TabPFN-2.5. Excess of target over blind quantifies the actual DI signal, accounting for distributional differences between $\sus$ and $\{\ho\}$.}
    \label{tab:internal-auc}
    \footnotesize
    \begin{tabular}{lllll}
    \toprule
    Target TFM $f_\theta$ & Classifier $g_\varphi$ & \multicolumn{1}{c}{Target TFM} & \multicolumn{1}{c}{Blind TFM} & \multicolumn{1}{c}{Excess}  \\ \midrule
    \multirow{3}{*}{SAP-RPT-1-OSS} & LR & 0.972 $\pm$ 0.001 & 0.714 $\pm$ 0.005 & $+$0.258 $\pm$ 0.005 \\
     & XGB & 0.992 $\pm$ 0.001 & 0.697 $\pm$ 0.004 & $+$0.295 $\pm$ 0.005 \\
     & TabPFN-2.5 & 0.994 $\pm$ 0.000 & 0.689 $\pm$ 0.005 & $+$0.306 $\pm$ 0.005 \\ \midrule
    \multirow{3}{*}{TabDPT} & LR & 0.675 $\pm$ 0.036 & 0.686 $\pm$ 0.035 & $-$0.011 $\pm$ 0.044 \\
     & XGB & 0.722 $\pm$ 0.034 & 0.662 $\pm$ 0.037 & $+$0.060 $\pm$ 0.036 \\
     & TabPFN-2.5 & 0.695 $\pm$ 0.035 & 0.611 $\pm$ 0.038 & $+$0.084 $\pm$ 0.034 \\ \midrule
    \multirow{3}{*}{nanoTabPFN} & LR & 0.510 $\pm$ 0.004 & 0.499 $\pm$ 0.004 & $+$0.011 $\pm$ 0.004 \\
     & XGB & 0.549 $\pm$ 0.004 & 0.503 $\pm$ 0.004 & $+$0.046 $\pm$ 0.005 \\
     & TabPFN-2.5 & 0.539 $\pm$ 0.004 & 0.497 $\pm$ 0.004 & $+$0.042 $\pm$ 0.006 \\ \midrule
    \multirow{3}{*}{Real-TabPFN-2.5} & LR & 0.548 $\pm$ 0.081 & 0.585 $\pm$ 0.079 & $-$0.037 $\pm$ 0.022 \\
     & XGB & 0.482 $\pm$ 0.074 & 0.528 $\pm$ 0.077 & $-$0.046 $\pm$ 0.050 \\
     & TabPFN-2.5 & 0.542 $\pm$ 0.080 & 0.560 $\pm$ 0.079 & $-$0.018 $\pm$ 0.043 \\
    \bottomrule
\end{tabular}
    \vspace{-10pt}
\end{table}

\begin{table}[t]
    \centering
    \setlength{\tabcolsep}{3pt}
\setlength{\extrarowheight}{-1pt}
\setlength{\aboverulesep}{0.5ex}
\setlength{\belowrulesep}{0.5ex}
\setlength{\cmidrulesep}{0.3ex}
    \small
    \caption{\textbf{Fingerprint Feature Contributions.} Mean absolute SHAP value of the top 10 features from $\bar S(\mathcal{D})$, computed for the SAP-RPT-1-OSS target TFM and TabPFN-2.5 blind TFM, with the XGB classifier. Larger values indicate that the feature contributes more to membership classification. Identifiers for signals, manipulations, and reductions reference Tables~\ref{tab:signals}, \ref{tab:manipulations}, and \ref{tab:reductions} in Appendix~\ref{sec:app-b}.}
    \label{tab:signal-shapley-sap}
    \begin{tabular}{lccccc}
    \toprule  Fingerprint Feature & Manipulation & Signal & Reduction & Target & Blind \\ 
    \midrule 
     TaskSwap\_Wasserstein\_Range & M10 & S14 & R9 & 0.8942 & 0.0000 \\
     TaskSwap\_KLD\_Range & M10 & S12 & R9 & 0.7583 & 0.0159 \\
     TaskSwap\_KLD\_Mean & M10 & S12 & R7 & 0.5271 & 0.1173 \\
     Retrieval\_KLD\_Mean & M19 & S12 & R7 & 0.3491 & 0.0046 \\
     TaskSwap\_Wasserstein\_Mean & M10 & S14 & R7 & 0.3368 & 0.0001 \\
     Retrieval\_KLD\_Std & M19 & S12 & R8 & 0.2926 & 0.0076 \\
     Retrieval\_Wasserstein\_Mean & M19 & S14 & R7 & 0.2662 & 0.0000 \\
     TaskSwap\_CELoss\_Mean & M10 & S5 & R7 & 0.1726 & 0.0043 \\
     TaskSwap\_KLD\_Std & M10 & S12 & R8 & 0.1500 & 0.0575 \\
     FeatureTransform\_Accuracy\_Range & M9 & S4 & R9 & 0.1480 & 0.1146 \\
\bottomrule
    \end{tabular}
\end{table}

\subsection{Which Features Are Most Revealing?}
\label{subsec:shapley}
\vspace{-4pt}
To identify which features within the fingerprints $\bar{S}(\mathcal{D})$ drive accurate membership prediction, we compute SHAP values \citep{shapley1953value, lundberg2017shap} for the classifier $g_\varphi$ used for internal auditing. \Cref{tab:signal-shapley-sap} shows the top 10 features with the greatest SHAP values computed for the XGB classifier trained on fingerprints from the SAP-RPT-1-OSS target TFM, as compared to the TabPFN-2.5 blind TFM (see \Cref{tab:signal-shapley-tabdpt} in Appendix~\ref{subsec:shapley_tabdpt} for TabDPT).
Two manipulations that change the task or the context, task swap (M10) and retrieval (M19), account for nine of the ten most important features, emphasizing the importance of the auditor having full control over TFM inputs. Among signals, two distributional measures dominate: the KL divergence (S12) and the Wasserstein distance (S14) between the predicted and the true label distribution.
Therefore, if the model outputs the correct \textit{label marginals} for a certain dataset, both distances are small. Notably, the blind baseline relies on a largely different set of features, as seen by the low correlation between contribution rankings. This provides further evidence that true membership signals are distinct from mere distributional signals.
\vspace{-6pt}
\subsection{What Drives Inference Success?}
\label{sec:success_factors}
\vspace{-4pt}
We noted in \Cref{sec:di-performance} that the default nanoTabPFN and Real-TabPFN-2.5 models did not leak significant membership information. These models differ from SAP-RPT-1-OSS and TabDPT in several key ways: nanoTabPFN is much smaller, less performant at tabular tasks, and was trained on a very large number (690k) of fixed synthetic datasets rather than real data; Real-TabPFN-2.5 was pre-trained on non-fixed synthetic data and only fine-tuned on real data using an L2 penalty with respect to the initial TabPFN-2.5 weights \citep{purucker2026private}. We find it likely that these factors entail that nanoTabPFN and Real-TabPFN-2.5 do not encode a meaningful amount of membership information. The alternative is that our DI method lacked discriminatory power to extract the membership information retained in these models, however we find this less likely given that our DI method successfully extracted information from two TFMs reaching up to 0.99 AUC.

On the other hand, we observed greater levels of leakage from SAP-RPT-1-OSS compared to TabDPT. While both models were entirely trained on real data, they too have important differences. SAP-RPT-1-OSS was trained on a corpus of 2.18M tables, while TabDPT used 123 (Appendix~\ref{sec:tfms-datasets}). To make up for this small corpus, TabDPT introduced an SSL-based task generation mechanism in which each time a dataset is used, the column labeled as $y$ is randomly swapped with one of the columns in $X$, while other columns are randomly shuffled or dropped. These manipulations disrupt the normal relationship between $X$ and $y$, potentially obfuscating membership information.

To isolate the reasons why some TFMs are more robust to membership leakage, we run ablations changing how nanoTabPFN is trained. Specifically, we: (a) reduce the number of pre-training datasets, so that each dataset has a relatively larger influence; (b) increase the model capacity, as a larger model could memorize more aspects of its training data; and (c) train on real instead of synthetic data.

\vspace{-2pt}
\textbf{Number of synthetic datasets.} We decrease the number of datasets in the pre-training corpus, while keeping the total number of gradient updates fixed to increase the exposure of each individual dataset to the model. Decreasing the number of synthetic datasets from $50,000$ to $100$ increases DI success only at the extreme: with $100$ synthetic member datasets we observe an increased ROCAUC of $0.72$. The synthetic datasets have relatively few rows ($\sim50$) and the loss plateaus quickly, with no overfitting occurring like in traditional learning methods, so only when extremely few datasets are used is there a measurable difference in membership leakage~(see \Cref{tab:nanotabpfn-data-ablation} in Appendix~\ref{app:ablations}).

\vspace{-2pt}
\textbf{Increased model capacity.} We increase nanoTabPFN's capacity by adding more layers ($\{6,8,18\}$), increasing embedding dimensionality~($\{192, 384, 768\}$), adding attention heads~($\{6,8,12\}$), and increasing the MLP hidden dimension~($\{768,1536,3072\}$). The smallest hyperparameter settings match the default nanoTabPFN parameters, while the largest hyperparameters match SAP-RPT-1-OSS, the most leaky model (although other training details still differ). Overall, we vary model size between $3.72$~M and $115.8$~M parameters. Even with nanoTabPFN at the same size as SAP-RPT-1-OSS, DI is not successful, see \Cref{tab:nanotabpfn-arch-ablation} in Appendix~\ref{app:ablations}.

\vspace{-2pt}
\textbf{Real vs. synthetic data.} We replace the fixed synthetic datasets drawn from a prior with real data from T4 \citep{gardner2024large}, matching a subset of the SAP-RPT-1-OSS dataset. When pre-trained with real data, ROCAUC of the DI reaches 0.811 with a single feature, and further increases to $0.92$ when all available features are used, see \Cref{tab:nanotabpfn-data-source} in Appendix~\ref{app:ablations}.
This shows that characteristics of real data not captured by synthetic priors lead to fingerprintable behavior differences in what the model learns.

\vspace{-2pt}
\textbf{Pre-train vs. FT on real data.} As a middle ground between real and prior-synthetic data, we use a generative model to create synthetic data based on the real datasets. In particular, we employ a Gaussian copula data synthesizer which fits an empirical cumulative distribution function (CDF) per column, scales the columns based on a standard normal, fits a multivariate Gaussian on the transformed columns, samples from that Gaussian, and inverts the CDFs to go back to the original column space. The result preserves marginals and pairwise correlations of the data. We fit one Gaussian copula for each real dataset and sample the same number of rows as each had for use in the DI method. Synthetic data generated from real datasets still yields a detectable membership signal (\Cref{tab:nanotabpfn-data-source} in Appendix~\ref{app:ablations}), suggesting that the signal is not solely tied to individual training rows, but can also arise from dataset-level statistical structure such as marginals and correlations.

%% file: content/6-conclusion.tex
In this work, we introduced dataset inference for tabular foundation models as a black-box auditing method for determining whether a suspect tabular dataset was part of a model’s pre-training corpus. By exploiting the in-context prediction interface of TFMs, we constructed dataset-level fingerprints from diverse input manipulations, output signals, and reductions, and showed that while individual classical membership signals are weak, a combination of TFM-specialized signals can reliably reveal dataset membership for TFMs trained on real tabular data, reaching near-perfect ROCAUC on SAP-RPT-1-OSS. Our ablations indicate that membership leakage is driven less by model capacity, and more by the use of real data in pre-training corpora. In other words, TFMs trained on fixed synthetic datasets are largely resistant to membership leakage except in the regime of extremely low numbers of distinct training datasets. Real data, and real-derived synthetic datasets, leave substantially stronger dataset-level traces that can be extracted by our auditing methods. These results suggest that dataset inference can serve as a practical tool for provenance and privacy auditing of deployed TFM APIs, helping data owners and model developers assess unauthorized data use and leakage risk. 

\section{Acknowledgments}
FB has received funding from the European Research Council (ERC) under the European Union’s Horizon Europe research and innovation programme (grant agreement No 101220235). This work was supported by the Helmholtz Association’s Initiative and Networking Fund on the HAICORE@FZJ partition.

%% file: content/D-appendix.tex
Our work has several limitations. First, successful dataset inference requires aggregating many weak features: individual signals such as loss or confidence provide only limited evidence of membership. Second, our method is mainly relevant for TFMs trained on fixed corpora of real or real-derived tabular datasets; many current TFMs, including the most performant ones, are trained on non-fixed synthetic data where there may be no meaningful finite training corpus to audit, and our experiments show little to no inference success on fixed synthetic data except in extreme low-diversity regimes. Third, our approach is model-agnostic and does not exploit architecture- or training-specific details, so stronger attacks may be possible when more information about the audited TFM is available.

%% file: content/A-appendix.tex
In \Cref{sec:tfms} we introduced the commonalities between existing TFMs, namely their ICL prediction mechanism. We also discussed the use of real and synthetic data for pre-training. Here, we give more detail on the diverse architectural details, data treatments, and training methods used for TFMs currently.

Internally, TFMs use an attention mechanism \citep{vaswani2017attention}, where predictions on $X^i_\text{qy}$ can attend to context datapoints, but not to other query datapoints. The actual implementation varies greatly, with some choices being to attend across rows (datapoints indexed by $i$) \citep{hollmann2023tabpfn, ma2025tabdpt}, across columns (elements of $X^i$), or some hybrid combination \citep{Hollmann2025, qu2025tabicl,  spinaci2025contexttab, bouadi2025orion}. These choices are further complicated due to the ways data can be encoded. Since tabular data is heterogeneous, where columns of $X$ could be floats, ints, dates, or categorical, among others, TFMs require an encoder as a first step to create a uniform vector representation. Choosing what and how to encode is a major design decision, but the most common choice is to tokenize entire rows \citep{hollmann2023tabpfn, ma2025tabdpt}, as opposed to individual cells.

To enhance the variety of training tasks, the prediction target $y$ can be randomly chosen out of all columns in the dataset, a form of self-supervised learning \citep{ma2025tabdpt, shaheen2026understanding}. This style of task creation has been successfully extended to foundation models for causal inference which also act on tabular data \citep{balazadeh2025causalpfn, stith2026causal}.

We also note a separate line of research focused on tabular prediction using LLMs. Due to the general capabilities of LLMs to reason and perform ICL, efforts have been made to adapt them for tabular prediction tasks through ICL \citep{jaitly2023towards, wen2024from}, or by post-training LLMs on tabular prediction tasks \citep{wang2023unipredict, gardner2024large, su2024tablegpt2, zhang2025tablellm, wen2025scalableincontextlearningtabular, wang2026tabsieve}. Because LLMs are pre-trained on vast quantities of non-tabular data and fundamentally process text tokens, we consider this setting to be distinct, and instead focus solely on TFMs.

%% file: content/B-appendix.tex
In this Appendix, we give complete details on how our dataset inference method is implemented for TFMs. As described in \Cref{sec:method}, from each dataset $\mathcal{D}$ we create a fingerprint vector $\bar{S}(\mathcal{D})$, where each element is a unique (manipulation, signal, reduction)-triple. These elements are used to train a model to perform anomaly detection of member datasets, or predict binary dataset member labels.

Signals are observable outputs from a TFM. In \Cref{tab:signals} we list the 14 signals constructed from the output of the TFM, which is the predictive distribution over labels $p(y^i_\text{qy} \mid X^i_\text{qy}; \mathcal{D}_\text{ctx})$. Signals include common measures for determining how well $f_\theta$ performs at the predictive task, like per-datapoint accuracy/error, predictive confidence, commonly used losses, and distributional measures like entropy and KL divergence. The KL and reverse-KL signals compare the output distribution $p(y^i_\text{qy} \mid X^i_\text{qy}; \mathcal{D}_\text{ctx})$ to a reference distribution $q(y)$. For concreteness, we use the distribution of true labels $y_\text{qy}$, although other options could be considered.

Manipulations are operations that can be performed on the inputs of $f_\theta$, namely $X_\text{qy}$ and $\mathcal{D}_\text{ctx}$. The auditor is free to construct these inputs from the dataset $\mathcal{D}$ (either $\sus$ or $\ho$) as they wish, which provides strong leverage for interrogating the TFM using only a black-box API. Manipulations are designed so that the predictive distribution will behave differently depending on if $\mathcal{D}$ is a member or non-member. In \Cref{tab:manipulations} we list the 20 different types of input manipulation we perform. Truth Serum is inspired by \cite{tramer2022truthserum}. Brainwash is inspired by \cite{wen2024brainwash}.

Finally, reductions first vary a parameter of the manipulation, observe a sequence of corresponding signals, and aggregate that sequence to create a feature, one of the elements of $\bar S(\mathcal{D})$ used for dataset inference downstream. Some reductions are designed to match properties of certain manipulations, like their intensity as applicable. In \Cref{tab:reductions} we detail the 11 types of reductions that we applied.

\begin{table}[t]
    \centering
    \footnotesize
        \caption{Output \textbf{signals} for dataset inference on TFMs. \textbf{C} - classification tasks; \textbf{R} - Regression tasks. $\hat y^i$ denotes the point prediction $\argmax_y p(y)$ for query row $i$.}
    \label{tab:signals}
    \setlength{\tabcolsep}{2pt}
\setlength{\extrarowheight}{-1pt}
\setlength{\aboverulesep}{0.5ex}
\setlength{\belowrulesep}{0.5ex}
\setlength{\cmidrulesep}{0.3ex}
    \begin{tabular}{llll}
    \toprule
        Signal Name & Type & Formula & Description \\
         \midrule
        \textbf{S1:} Correct class confidence & \textbf{C} & $p(y):=p(y \mid  X^i_\text{qy}; \mathcal{D}_\text{ctx})$ & Likelihood assigned to the correct class \\
        \textbf{S2:} Max confidence & \textbf{C}& $\max_y p(y)$ & Highest likelihood assigned to any label \\
        \textbf{S3:} Prediction margin & \textbf{C} & $p(y_{(1)}) - p(y_{(2)})$ & Difference of highest two class likelihoods \\
        \textbf{S4:} Accuracy & \textbf{C} & $\mathds{1}(\hat y^i = y^i_{\text{qy}})$ & Correctness of predicted class \\
        \textbf{S5:} Cross entropy & \textbf{C} & $-\sum_{k=1}^{K} y^i_\text{qy}\log {p}_{k}(y)$ & Cross entropy loss\\
        \textbf{S6:} Absolute error & \textbf{R} & $\vert \hat y^i - y^i_{\text{qy}}\vert$ & Absolute error loss \\
        \textbf{S7:} Mean squared error & \textbf{R} & $(\hat y^i - y^i_{\text{qy}})^2$
 & Mean squared error loss \\
        \textbf{S8:} R$^2$ & \textbf{R} & $1 - {\sum_i (y_i - \hat{y}_i)^2}/{\sum_i (y_i - \bar{y})^2}$ & Coefficient of Determination \\
        \textbf{S9:} Prediction Spread & \textbf{R} & $\sigma(\hat{y}) / \sigma(y)$ & Relative standard deviation \\
        \textbf{S10:} Chi-Squared$^\dagger$ & \textbf{CR} & $\sum_{k=1}^K (p(y)_k- q(y)_k)^2/q(y)_k$ & Comparison of class frequencies to $q(y)$ \\
        \textbf{S11:} Entropy & \textbf{CR} & $-\sum_y p(y) \log p(y)$ & Shannon entropy of class probabilities \\
        \textbf{S12:} KL Divergence & \textbf{CR} & $\sum_y p(y)\log\frac{p(y)}{q(y)}$ & KL Divergence to a distribution $q(y)$\\
        \textbf{S13:} Reverse KL & \textbf{CR} & $\sum_y q(y)\log\frac{q(y)}{p(y)}$ & KL Divergence from a distribution $q(y)$\\
        \textbf{S14:} Wasserstein Distance & \textbf{CR} & $W_1(\hat{y}, y)$ & Wasserstein-1 to reference output $y$ \\
        \bottomrule
    \end{tabular}
    \scriptsize  $\dagger$: Computed in 30 bins for \textbf{R}.
\end{table}

\begin{table}[t]
    \centering
    \small
    \setlength{\tabcolsep}{3pt}
\setlength{\extrarowheight}{2pt}
\setlength{\aboverulesep}{0.5ex}
\setlength{\belowrulesep}{0.5ex}
\setlength{\cmidrulesep}{0.3ex}
        \caption{Input \textbf{manipulations} for dataset inference on TFMs and their parameters. \textbf{C} - classification tasks; \textbf{R} - Regression tasks}
    \label{tab:manipulations}
    \begin{tabular}{ll p{6cm} p{3.5cm}}
    \toprule
        Manipulation Name & Type & Description & Parameters\\
         \midrule
        \textbf{M1:} No manipulation & \textbf{CR} & Only split rows of $\mathcal{D}_\text{sus}$ into disjoint sets $(X_\text{ctx}, y_\text{ctx})$ and $(X_\text{qy}, y_\text{qy})$ & Splitting fraction (applies to all below)\\
        \hline
        \textbf{M2:} Query = Context & \textbf{CR} & Repeat the context inputs $X_\text{ctx}$ as $X_\text{qy}$ & - \\
                \hline
        \textbf{M3:} Query only & \textbf{CR} & Provide only $X_\text{qy}$, with $\mathcal{D}_\text{ctx}=\varnothing$ & - \\
                \hline
        \textbf{M4:} Row sampling & \textbf{CR} & Subsample the rows of $\mathcal{D}_\text{ctx}$ & Fraction of rows sampled\\
                \hline
        \textbf{M5:} Row shuffling & \textbf{CR} & Randomly permute the rows of $\mathcal{D}_\text{ctx}$ & Permutation\\
                \hline
        \textbf{M6:} Feature sampling & \textbf{CR} & Subsample the columns of $X_\text{ctx}$ & Fraction of columns sampled \\
                \hline
        \textbf{M7:} Feature shuffling & \textbf{CR} & Randomly permute the columns of $X_\text{ctx}$ & Permutation \\
                \hline
        \textbf{M8:} Feature noising & \textbf{CR} & Randomly perturb the columns of $\tilde{X}_\text{ctx}{=}X_\text{ctx}{+}\varepsilon$ & Noise type (e.g. Gaussian); Noise level\\
                \hline
        \textbf{M9:} Feature transform & \textbf{CR} & Transform a column ($\mathcal{D}_\text{ctx}$ and $\mathcal{D}_\text{qy}$) with a fixed function & Column; Function (e.g. $\log$)\\
                \hline
        \textbf{M10:} Task swap & \textbf{CR} & Redefine the label $y$ as one of the columns of $X$ & Column\\
                \hline
        \textbf{M11:} Class permutation & \textbf{C} & Re-encode classes as $y'=\pi(y)$ with a permutation $\pi$ & Permutation\\
                \hline
        \textbf{M12:} Label corruption & \textbf{CR} & Randomly select context rows and change $y_\text{ctx}$ to an incorrect label & Fraction corrupted\\
                \hline
        \textbf{M13:} Constant labels & \textbf{CR} & Set all context labels to a fixed value $y_\text{ctx} = a$ & Value\\
                \hline
        \textbf{M14:} Row duplication & \textbf{CR} & Randomly duplicate some rows of $\mathcal{D}_\text{ctx}$ & Fraction duplicated\\
                \hline
        \textbf{M15:} Truth serum & \textbf{CR} & Randomly duplicate some rows of $\mathcal{D}_\text{ctx}$, but with corrupted $y_\text{ctx}$ & Fraction duplicated\\
                \hline
        \textbf{M16:} Query leakage & \textbf{CR}  & Randomly copy some rows of $\mathcal{D}_\text{qy}$ into  $\mathcal{D}_\text{ctx}$ & Fraction copied\\
                \hline
        \textbf{M17:} Brainwash & \textbf{CR} & Randomly copy some rows of $\mathcal{D}_\text{qy}$ into  $\mathcal{D}_\text{ctx}$, but with corrupted $y_\text{ctx}$ & Fraction copied\\
                \hline
        \textbf{M18:} Nearest neighbours & \textbf{CR} & Compute NN within $\mathcal{D}$, and split them across $\mathcal{D}_\text{ctx}$ and $\mathcal{D}_\text{qy}$ & Number of neighbours \\
                \hline
        \textbf{M19:} Retrieval & \textbf{CR} & Retrieve NN of $X_\text{qy}$ from $\mathcal{D}$ as $X_\text{ctx}$ & Number of neighbours\\
                \hline
        \textbf{M20:} Softmax temperature & \textbf{C} & Divide the output logits by a temperature $T$ before the softmax & Temperature $T$\\
        \bottomrule
    \end{tabular}
\end{table}

\begin{table}[h]
    \centering
    \small
    \caption{\textbf{Reductions} vary the parameters of a manipulation to create a sequence of signals that are then summarised into a scalar entry of $\bar{S}(\mathcal{D})$. The applicable family depends on the structure of the manipulation's parameters: \emph{ordered} along an intensity axis, \emph{unordered}, or based on prediction \emph{stability}.}
    \label{tab:reductions}
    \setlength{\tabcolsep}{2pt}
    \begin{tabular}{ll p{5.5cm} p{4.5cm}}
    \toprule
        Type & Reduction & Formula & Captures \\
        \midrule        
        Default & \textbf{R1:} No Reduction & Signal is used directly as feature & Predictive power of signal directly\\
        \midrule
        Ordered & \textbf{R2:} Slope & Regression coef.\ on (log-spaced) intensity & Rate of signal change as manipulation intensity increases \\
        & \textbf{R3:} AUC & Normalised trapezoidal integral & Average signal value across manipulation intensity range \\
        & \textbf{R4:} Gain & Final value minus initial value & Total signal change between highest and lowest intensity \\
        & \textbf{R5:} Early & Value at 25\% intensity, normalised by max & Signal response at low-intensity manipulation \\
        & \textbf{R6:} Late gain & Final value minus second-to-last & End-of-curve flattening or runaway signal \\
        \midrule
        Unordered & \textbf{R7:} Mean & Mean over parameter range & Typical signal value \\
        & \textbf{R8:} St. dev. & Sample st. deviation over parameter range & Signal spread \\
        & \textbf{R9:} Range & Max minus min over parameter range & Signal range \\
        \midrule
        Stability & \textbf{R10:} Consistency & Pairwise signal agreement & Cross-parameter signal stability \\
        & \textbf{R11:} Rank corr & Mean pairwise Spearman correlation across per-row signals & Cross-parameter ranking stability \\
        \bottomrule
    \end{tabular}
\end{table}

%% file: content/C-appendix.tex
\subsection{Tabular Foundation Models and Datasets}\label{sec:tfms-datasets}
 Our DI experiments used four TFMs: TabDPT \citep{ma2025tabdpt} (Apache 2.0 license), SAP-RPT-1-OSS \citep{spinaci2025contexttab} (Apache 2.0 license), nanoTabPFN \citep{pfefferle2025nanotabpfn} (Apache 2.0 license via TFM-Playground \citep{tfm_playground_2025}), and Real-TabPFN-2.5 \citep{grinsztajn2025tabpfn} (TABPFN-2.5 License v1.1, which allows our research uses). Each of these models was trained on a different set of datasets. We describe the member datasets for each TFM below.

 \textbf{TabDPT} \citep{ma2025tabdpt} was trained on a collection of 123 datasets from OpenML \citep{vanschoren2014openml} (CC-BY license). For held-out non-members we use CC18 \citep{bischl2021cc18} and CTR23 \citep{fischer2023ctr}, also from OpenML, which TabDPT used as a test set, not for pre-training.

 \textbf{SAP-RPT-1-OSS} \citep{spinaci2025contexttab} was trained on a subset of the T4 dataset \citep{gardner2024large} (available under a custom license). This subset first excluded any dataset with 150 or fewer rows, leaving 2.18M tables available from pre-training. As this amount of tables is excessively large for testing DI, we select tables in chunk-002 from T4 (with >150 rows) as the members. The datasets with 150 or fewer rows are known non-members. Given this setup, we note that DI is trivial if row count is used as a predictive feature. Hence, when performing DI we ensure any feature related to row count or table size is excluded, and also use blind baselines to account for the performance level that can be obtained from distributional differences alone.

 \textbf{nanoTabPFN} \citep{pfefferle2025nanotabpfn} and its pre-training datasets are available in the TFM-Playground repository \citep{tfm_playground_2025} (Apache 2.0 license). Using default settings, this model is trained on synthetic data from a prior, but unlike other TFMs \citep{grinsztajn2025tabpfn} which generate new synthetic data for each batch, nanoTabPFN trains on a fixed set of synthetic tables that are recycled each epoch. The fixed set consists of 690k generated tables as members, and another 690k as non-members.

\textbf{Real-TabPFN-2.5}
\citep{grinsztajn2025tabpfn} starts from the pre-trained TabPFN-2.5 model, which was trained entirely on synthetic datasets drawn fresh from a prior each batch, and fine-tunes on a set of 43 real tables from OpenML and Kaggle. We use the 20 datasets from OpenML as training members, and include 51 datasets from TabArena \citep{erickson2025tabarena} (originally from OpenML) as non-members, since Real-TabPFN-2.5 used them only for testing. We note a few relevant points about how Real-TabPFN-2.5 was fine-tuned which were not disclosed in the original paper: fine-tuning lasted 25,000 gradient steps and did not use gradient aggregation; the loss included an L2 penalty with respect to the starting weights of TabPFN-2.5 \citep{purucker2026private}.

 In comparison, many other TFMs are not trained on a fixed set of datasets, but instead draw new synthetic datasets from a prior for each batch during pre-training. Such synthetic datasets generally do not come with privacy or provenance concerns, so we do not consider DI on these TFMs. However, this fact also makes them suitable as blind baselines. In particular, we use the highly-performant \textbf{TabPFN-2.5} \citep{grinsztajn2025tabpfn}, with model weights released by the authors under the custom TABPFN-2.5 License v1.1.

\subsection{Single Feature Dataset Inference on TabDPT}
\label{sec:app-signal-separation-tabdpt}

\begin{figure}[t]
  \centering
  \begin{subfigure}[b]{0.32\linewidth}
    \includegraphics[width=\linewidth]{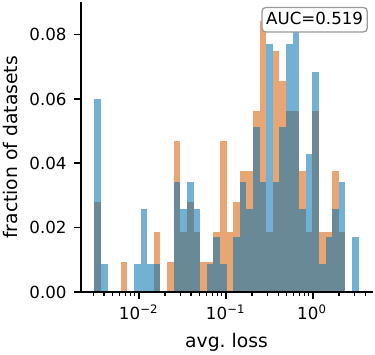}
  \end{subfigure}\hfill
  \begin{subfigure}[b]{0.32\linewidth}
    \includegraphics[width=\linewidth]{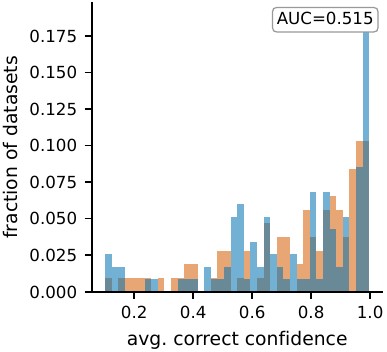}  \end{subfigure}\hfill
  \begin{subfigure}[b]{0.32\linewidth}
    \includegraphics[width=\linewidth]{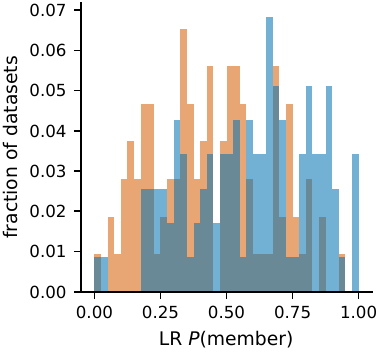}
  \end{subfigure}\\[0.3em]
  \includegraphics[width=0.3\linewidth]{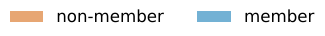}
  \caption{\textbf{Single- and multi-feature dataset inference on TabDPT.} The \textit{per-dataset} average loss (Left) and correct-class confidence (Middle) computed for member and non-member datasets of TabDPT \citep{ma2025tabdpt} ($n{=}224$ datasets) do not clearly separate member and non-member datasets, as in \Cref{fig:signal-separation-single}. (Right) The internal auditing method used as in \Cref{fig:signal_separation} with many TFM-specific features achieves higher ROCAUC.}
  \label{fig:signal-separation-tabdpt}
\end{figure}

\Cref{fig:signal-separation-tabdpt} reproduces \Cref{fig:signal-separation-single} on TabDPT ($n{=}224$ datasets), showing the strength of individual features for dataset inference. Single-signal AUCs are again close to chance ($0.61$ for both average loss and average correct confidence).

\subsection{Feature Contributions}\label{subsec:shapley_tabdpt}
\Cref{tab:signal-shapley-sap} in \Cref{subsec:shapley} showed the top 10 features in $\bar S(\mathcal{D})$ for the classifier $g_\varphi$ for internal auditing, using SAP-RPT-1-OSS as the target TFM. In \Cref{tab:signal-shapley-tabdpt} we show the same analysis using TabDPT as the target TFM, with all other settings unchanged. Like for \Cref{tab:signal-shapley-sap}, dataset inference on the blind TFM relies on an entirely different set of features than for the target TFM. Compared to \Cref{tab:signal-shapley-sap}, inference on TabDPT has heaviest reliance on a different set of manipulations, signals, and reductions. However, the auditor need not know the identity of the TFM they are querying for dataset inference to work; our tests in \Cref{tab:internal-auc} are model agnostic and generate a large number of features in $\bar S(\mathcal{D})$, relying on the classifier $g_\varphi$ to automatically choose which features are predictive. Strong classifiers like XGBoost are fully capable of filtering out noisy features when appropriate regularization is used.

\begin{table}[t]
\centering
    \setlength{\tabcolsep}{3pt}
\setlength{\extrarowheight}{-1pt}
\setlength{\aboverulesep}{0.5ex}
\setlength{\belowrulesep}{0.5ex}
\setlength{\cmidrulesep}{0.3ex}
    \caption{\textbf{Fingerprint Feature Contributions.} Mean absolute SHAP value of the top 10 features from $\bar S(\mathcal{D})$, computed for the TabDPT target TFM and TabPFN-2.5 blind TFM, with the XGB classifier. Larger values indicate that the feature contributes more to membership classification. Identifiers for signals, manipulations, and reductions reference Tables~\ref{tab:signals}, \ref{tab:manipulations}, and \ref{tab:reductions} in Appendix~\ref{sec:app-b}.}
    \label{tab:signal-shapley-tabdpt}
    \begin{tabular}{lccccc}
    \toprule
    Fingerprint Feature & Manipulation & Signal & Reduction & Target & Blind \\
    \midrule
     RowSampling\_KLDClean\_LateGain & M4 & S12 & R6 & 0.2203 & 0.0559 \\
     FeatureNoise\_KLD\_Early & M8 & S12 & R5 & 0.1710 & 0.0074 \\
     TruthSerum\_Confidence\_Early & M15 & S1 & R5 & 0.1573 & 0.0359 \\
     RowDuplication\_ChiSquared\_Early & M14 & S10 & R5 & 0.1523 & 0.0077 \\
     FeatureSample\_KLD\_AUC & M6 & S12 & R3 & 0.1504 & 0.0747 \\
     FeatureSample\_KLD\_Slope & M6 & S12 & R2 & 0.1427 & 0.0885 \\
     TruthSerum\_CELoss\_LateGain & M15 & S5 & R6 & 0.1307 & 0.0057 \\
     TruthSerum\_KLD\_Early & M15 & S12 & R5 & 0.1271 & 0.0136 \\
     Temperature\_KLD\_Mean & M20 & S12 & R7 & 0.1008 & 0.0097 \\
     SplitFraction\_Wasserstein\_Gain & M1 & S14 & R4 & 0.0864 & 0.0017 \\
    \bottomrule
    \end{tabular}
\end{table}

\subsection{Ablations}
\label{app:ablations}
We perform ablations on nanoTabPFN to determine under which conditions it leaks membership information under our DI method. We use nanoTabPFN as it is modularly designed and relatively lightweight to train, finishing a training run in a few minutes on a single H100 as \Cref{tab:nanotabpfn-training-time} shows. We ablate over architecture size: we interpolate model size between the default size and the $32\times$ larger SAP-RPT-1-OSS configuration. First we reduce the number of datasets. \Cref{tab:nanotabpfn-data-ablation} shows that only with very few datasets ($100$) do we see a significant increase in AUC. \Cref{tab:nanotabpfn-arch-ablation} shows that increased model size does not lead to membership leakage. Finally, we train nanoTabPFN on real data, and \Cref{tab:nanotabpfn-data-source} shows this allows differentiating member and non-member data. We also derive synthetic data from the real data using a Gaussian Copula generator. This barely reduces AUC from $0.92$ for real data to $0.90$ for the real data-derived synthetic data. 

\begin{table}[t]
  \centering
  \small
  \caption{\textbf{NanoTabPFN training cost.} Single-GPU (NVIDIA H100) training time
    for the default checkpoint and all ablation variants used in the paper.
    Architecture is given as L\,/\,E\,/\,H\,/\,M (layers, embedding dim,
    attention heads, MLP hidden dim). All variants are trained for 80 epochs
    of 25 steps each at batch size 50 with AdamW ($\eta = 10^{-4}$).}
  \label{tab:nanotabpfn-training-time}
  \begin{tabular}{lrrrr}
    \toprule
    Variant       & L/E/H/M               & Prior tables & Time/epoch & Compute \\
    \midrule
    \texttt{default}     & 6/192/6/768   & 50\,000  & 0.55\,s & 44\,s   \\
    \texttt{data25k}     & 6/192/6/768   & 25\,000  & 0.54\,s & 43\,s   \\
    \texttt{data10k}     & 6/192/6/768   & 10\,000  & 0.52\,s & 42\,s   \\
    \texttt{data5k}      & 6/192/6/768   &  5\,000  & 0.52\,s & 42\,s   \\
    \texttt{arch\_medium}& 8/384/8/1536  & 50\,000  & 1.34\,s & 1.8\,min \\
    \texttt{arch\_large} & 12/768/12/3072& 50\,000  & 5.35\,s & 7.1\,min \\
    \bottomrule
  \end{tabular}
\end{table}

\begin{table}[t]
  \centering
  \caption{\textbf{NanoTabPFN training corpus size ablation.} Dataset inference results for nanoTabPFN models trained on different sized corpora of prior-synthetic tables. For each row, the number of training steps and batch size is held constant, meaning that when the corpus is smaller, the model is exposed to each dataset more times. DI is run on a held-out synthetic eval set with equal numbers of member and non-member tables. Top-1 is the best single-features AUC; multivariate is logistic regression on the top-50 features.}
  \label{tab:nanotabpfn-data-ablation}
  \begin{tabular}{rrccc}
    \toprule
    $|\mathcal{D}_{\text{train}}|$ & eval $n$ & top-1 AUC & multivariate AUC \\
    \midrule
    50,000 & 100,000 & 0.505 & $0.501 \pm 0.003$ \\
      25,000 & 50,000 & 0.511 & $0.506 \pm 0.004$ \\
  10,000 & 20,000 & 0.516 & $0.514 \pm 0.005$ \\
  5,000 & 10,000 & 0.531 & $0.523 \pm 0.009$ \\
    2,000 & 4,000 & 0.530 & $0.516 \pm 0.019$ \\
    1,000 & 2,000 & 0.536 & $0.541 \pm 0.037$ \\
  500 & 1,000 & 0.570 & $0.563 \pm 0.037$ \\
    100 & 200 & 0.632 & $0.720 \pm 0.061$ \\
    \bottomrule
  \end{tabular}
\end{table}

\begin{table}[t]
  \centering
  \caption{\textbf{NanoTabPFN architecture size ablation.} All variants are trained on the same 50\,000 synthetic prior tables; only the parameter count differs. ``L/E/H/M'' denotes layers, embedding dimension, attention heads, and feed-forward dimension. DI evaluation is run on the same held-out 50\,000 member / 50\,000 non-member split as in \Cref{tab:nanotabpfn-data-ablation}.}
  \label{tab:nanotabpfn-arch-ablation}
  \begin{tabular}{llccc}
    \toprule
    Variant & L/E/H/M & params & top-1 AUC & multivariate AUC \\
    \midrule
  Small (default) & 6/192/6/768 & 3.7M &  0.505 & $0.501 \pm 0.003$ \\
  Medium & 8/384/8/1536 & 19.5M &  0.506 & $0.505 \pm 0.003$ \\
  Large & 12/768/12/3072 & 115.8M & 0.505 & $0.503 \pm 0.004$ \\
    \bottomrule
  \end{tabular}
\end{table}

\begin{table}[t]
  \centering
  \caption{\textbf{NanoTabPFN training-corpus ablation.} DI results on two nanoTabPFN models with the default architecture, trained on the real T4 corpus and on a Gaussian-copula synthetic copy of T4.}
  \label{tab:nanotabpfn-data-source}
  \begin{tabular}{lcccc}
    \toprule
    Training corpus & members & non-mem.\ & top-1 ROCAUC & multivariate ROCAUC \\
    \midrule
  Real T4 & 9,321 & 3,725 & 0.811 & $0.923 \pm 0.006$ \\
  Synthetic, Gaussian copula & 8,867 & 2,437 & 0.871 & $0.897 \pm 0.008$ \\
    \bottomrule
  \end{tabular}
\end{table}